# InstaGrasp: An Entirely 3D Printed Adaptive Gripper with TPU Soft Elements and Minimal Assembly Time*

Xin Zhou, *Student Member, IEEE*, and Adam J. Spiers, *Member, IEEE*

*Abstract*—Fabricating existing and popular open-source adaptive robotic grippers commonly involves using multiple professional machines, purchasing a wide range of parts, and tedious, time-consuming assembly processes. This poses a significant barrier to entry for some robotics researchers and drives others to opt for expensive commercial alternatives. To provide both parties with an easier and cheaper (under £100) solution, we propose a novel adaptive gripper design where every component (with the exception of actuators and the screws that come packaged with them) can be fabricated on a hobby-grade 3D printer, via a combination of inexpensive and readily available PLA and TPU filaments. This approach means that the gripper's tendons, flexure joints and finger pads are now printed, as a replacement for traditional string-tendons and molded urethane flexures / pads. A push-fit systems results in an assembly time of under 10 minutes. The gripper design is also highly modular and requires only a few minutes to replace any part, leading to extremely user-friendly maintenance and part modifications. An extensive stress test has shown a level of durability more than suitable for research, whilst grasping experiments (with perturbations) using items from the YCB object set has also proven its mechanical adaptability to be highly satisfactory.

## I. Introduction

In recent years, there has been an explosion in number of available robotic hands. 2019 alone has seen more than 200 published hands – an increase of around 35% from just two years before [1]. The increasing popularity of this field is mainly driven by a rising availability of 3D printing technologies and open-source initiatives such as the Yale OpenHand Project [2].

A large number of gripper prototypes now rely on 3d printing for the majority of their components. This includes all designs under the OpenHand Project umbrella (e.g. Model T42 [3], Model VF [4]), as well as many others such as the Pisa/IIT Softhand [5], CLASH [6], and the E-TRoll hand [7].

Although these hands are significantly cheaper to acquire and easier to modify than many commercially available grippers, fabricating, maintaining, and modifying them is still a large hurdle for robotics researchers, particularly if they have little background in mechanical engineering. To demonstrate this point: The Yale OpenHand Model T42 [3], which is one of the simplest and most popular hands in the OpenHand family, uses over 10 different types of non-printable parts (bolts, heat-set inserts, dowel pins, tendon strings etc.) and requires fabrication processes such as urethane casting, degassing, and precise hole drilling (ideally with a drill press and reamer tools). Tendon routing and tensioning is a further step. It is also notable that the part geometries for this hand as well as many others require the use of higher-end 3D printers for a clean print (e.g., dual nozzle 3D printers with soluble supports). For instance, the OpenHand prototypes typically use parts printed on a Stratsys Fortus, which retails at over £10K. It is not surprising that purchasing an expensive commercial gripper is still the path of lowest effort for many researchers. For context, the Robotiq 2F-85, which is one of the best-selling grippers, sold for £2,640 in 2022.

* Research supported by Imperial College London internal funds.
Both authors are with the Manipulation and Touch Lab, Department of Electrical and Electronic Engineering, Imperial College London, UK (corresponding author: xin.zhou16@imperial.ac.uk).

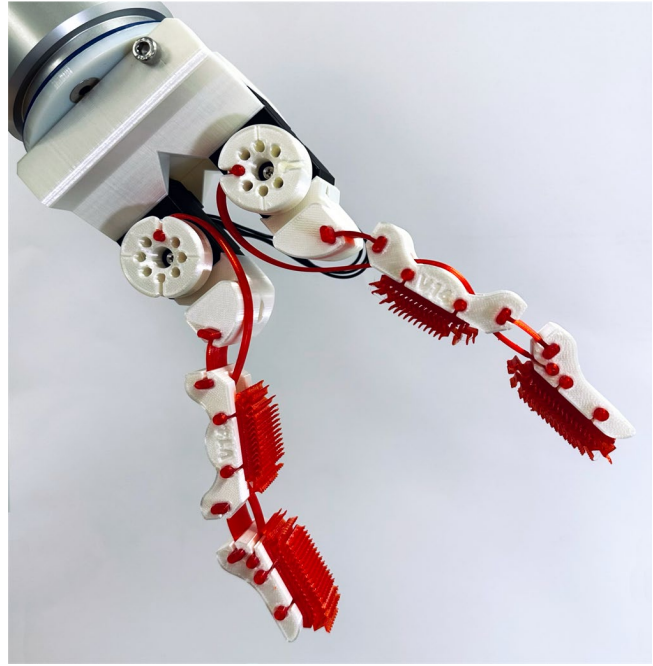

Figure 1: The underactuated InstaGrasp features rigid and flexible components that are 3D printable on hobby grade machines. The only screws used are packaged with the actuators. All parts apart from actuators and those screws are 3D printed.

Hence, this work started with one clear goal in mind: creating a low-cost adaptive robotic gripper design pushing the ease of fabrication, maintenance, and modifications to their maximum. The resulting gripper design is named InstaGrasp (Fig. 1), and only utilizes two types of non-printable parts: two low-cost Dynamixel XL430-W250-T servo motors (£41 each) and the screws that are *already included* in the servo package. The higher performance (and cost) Dynamixel XM430 servomotors are also directly compatible with the InstaGrasp housing, if larger gripping torque is required.

TABLE I. PARTS LIST MODEL T42* VS INSTAGRASP

|  | InstaGrasp | Model T42* |
|---|---|---|
| Fabrication Equipment | Hobby 3D Printer. | Professional 3D Printer (e.g., dual nozzle printer or SLS printer), Vacuum Chamber, Power Drill, Belt Sander (Optional), Precision Reamer Bits (Optional). |
| 3D Printed Parts | 6 PLA Finger Parts, 2 PLA Servo Pulleys, 1 PLA Base, 2 TPU Tendons, 4 TPU Joints, 4 TPU Finger Pads. | 7 Finger Assembly Parts, 2 Pulleys, 6 Base Parts, 2 Optional Covers. |
| Purchased Parts | 2 Dynamixel Servos (Includes All Required Screws), 1 Spool of PLA, 1 Spool of TPU. | 2 Dynamixel Servos, 1 Spool of Printing Material, 2 Types of Urethane, 1 Spool of Tendon, 4 Heat-Set Inserts, 4 Female Standoffs, 4 Metal Dowel Pins, 4 Additional Types of Screws/Bolts, 1 Fan. |

*Flexure-flexure Model T42 version 1.0 using Dynamixel servos.

The InstaGrasp is based on the successful finger configuration of the T42 gripper, yet is significantly easier to fabricate. Table 1 compares parts needed to fabricate an InstaGrasp vs a Model T42, highlighting the InstaGrasp's ease of assembly as well as effort saved in not having to acquire numerous specific parts.

Only a hobby grade 3D printer (such as a £750 Prusa i3) is required to fabricate all mechanical parts, including flexures and flexible tendons, which are printed in flexible thermoplastic polyurethane (TPU). All part designs are optimized for single nozzle 3d printers with default hardware and print settings. Even print support structures are avoided to eliminate the need for post-processing of parts. The flexible elements are based on low-cost and widely available 'Amazon Basics' TPU filament from Amazon.co.uk (£18 for 1 kg), whilst PLA (Polylactic acid) filament from the same brand (£17 for 1 kg) was used for the rigid parts.

Instead of relying on screws or other fasteners, the InstaGrasp is held together via a push-fit system: All TPU parts have "pins" that slot into the PLA parts. These TPU pins expand inside the PLA slots and keep parts tightly fastened. Assembly and part replacements are then simply via push-fit, requiring only a hammer for insertion and combination pliers or flat-nose pliers for removal.

As a result, one can fabricate the InstaGrasp in 3 easy steps:

1. Purchase two Dynamixel servo motors and 3D printing filament (PLA and TPU) if needed.

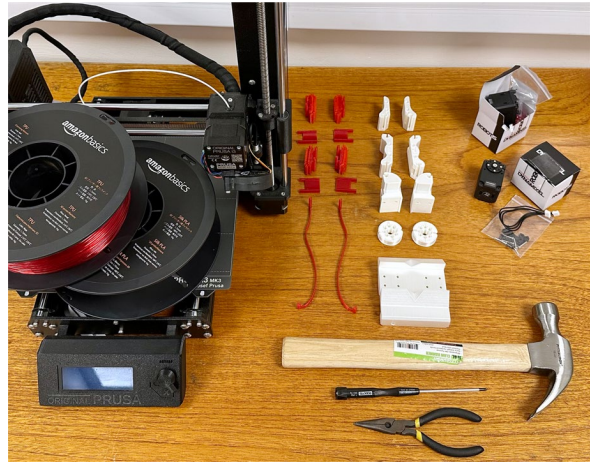

Figure 2: All equipment, tools, and parts needed to fabricate the InstaGrasp are shown here. The TPU (center top in red) and PLA parts (center top in white) are printed on the Prusa i3 MK3S (left). All required screws are included in the Dynamxiel servo packages (top right), and only a hammer, pliers, and a screwdriver (bottom right) are needed for assembly and part replacements.

2. Print 9 PLA parts and 10 TPU parts, which can be done on a hobby-grade printer with no print supports. (£2.50 material cost, 17.5 hours printing time)

3. Assemble in under 10 minutes.

Fig. 2 presents all equipment, tools and parts needed to fabricate and modify an InstaGrasp.

For the T42, we estimate the part fabrication (3D printing and urethane curing) to take around 30 hours, and part processing (including drilling, sanding, moulding preparations and cleanup, rubber pouring, assembling, etc.) to take at least 2-3 hours. Clearly, the T42 as well as other hands have numerous different parts and processes in order to improve performance and reliability. With a processing time of under 10 minutes, the InstaGrasp does not aim to outcompete these hands. For example, the Model T42's rubber casted finger pads are bound to have a much higher friction due to the urethane's significantly higher softness compared to the TPU. However, as will be clear in later sections, the unique geometry of the TPU finger pads perform adequately in grasping tasks. Therefore, just like the full InstaGrasp, they can serve as a great low effort starting point for grasping tasks and rapid prototype iterations. If more grip is needed, researchers can easily add on their custom-made finger pads via finger pad adapters, which will be showcased in later sections as well.

The following sections of this paper will outline the system design, present the surprisingly high durability of the 3D printed TPU flexures and tendons, and demonstrate the grasping performance of the InstaGrasp. Considering its total cost of under £100 and extremely quick and easy assembly, we believe that this gripper design can close the gap between expensive commercial grippers and existing open-source grippers with high fabrication overhead. The ability to easily modify the InstaGrasp will also allow other researchers to rapidly design and test new component variations for specific tasks or general performance improvement. We encourage such modifications to be shared with the community.

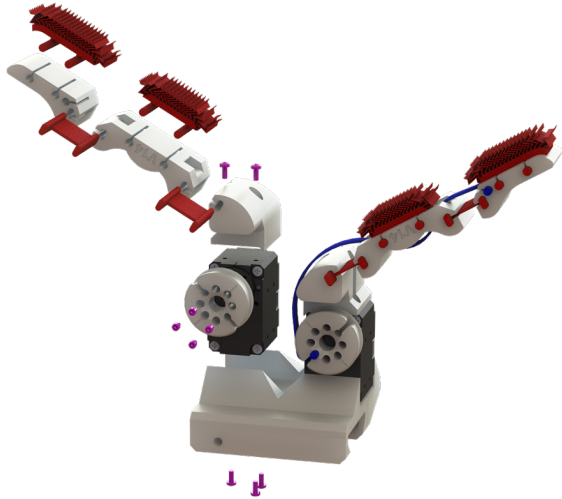

Figure 3: Exploded view of the InstaGrasp, showing TPU flexure joints and TPU finger pads in red, TPU tendon in blue, and the screws (included in the servo purchase) in purple.

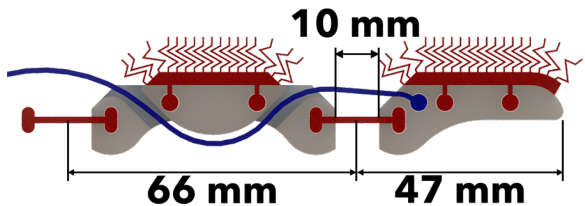

Figure 4: Cross section of the adaptive two-link finger. The tendon is shown in blue and connected to the pulley on the left side outside this image. Flexure joints have a moment arm of around 15 mm. Finger links are 30 mm wide (in the direction pointing into the page).

## II. HAND DESIGN

### A. Hand Configuration and Fingers

An exploded view of the InstaGrasp is given in Fig. 3, presenting the very few parts required to fabricate this hand. The overall configuration of this underactuated hand is largely aligned with the flexure-flexure version 1.0 of the Yale OpenHand Model T42 [3], sharing similar finger dimensions as well as tendon-routing approaches (Fig. 4). The InstaGrasp does not make use of metal dowel pins to act as a low-friction string-tendon pulleys, instead using curved features in the rigid parts to allow a larger bending radius for the relatively stiff TPU tendon. The base of the two fingers are roughly 70 millimeters apart, around 18mm wider than the T42 to avoid tendon clashing. Both flexure joints and servo pulley have a moment arm of around 15 millimeters. An unmodified Prusa i3 MK3S printer is used in this work to print all 3D printed parts, adding up to a total printing time of around 17.5 hours. Although Dynamixel XL430-W250 servo motors were used in this prototype, the base allows for the more powerful XM430-W350 servos to be mounted, which would almost triple the stall torque (4.1 N·m versus 1.4 N·m).

### B. Easy-Release Push-Fit System

One of the novel features of this hand is the TPU-based push-fit system, largely replacing the need for screws and bolts (Fig. 5). Apart from the significantly increased ease of assembly, an additional benefit is a much more modular design compared to the T42. This approach results in extremely easy replacements of individual joints, finger links, and finger pads following failure, or to test part modifications. In comparison, the T42 has these finger components permanently fused with cast urethane. All these TPU parts have 3 mm of the fitting pins protrude by design (Fig.4 bottom) and can be easily removed via pulling on them with a pair of flat-nose pliers. Part replacements can then be performed in mere minutes.

### C. TPU Tendons, Joints, and Finger Pads

Most tendon-driven hands utilize thin tendon strings or cables to actuate fingers such as in [3]–[6], [8], which comes with a multitude of benefits: these tendons are extremely bendable, which is great for flexible tendon routing, and do not easily stretch allowing reliable and efficient actuation. In this work however, we decided to explore the possibility of 3D printing tendons using flexible TPU (Fig. 5). Combined with our easy-release push-fit system described above, this design choice eliminates the need for tendon tensioning, fastening, and purchasing in addition to drilling routing holes accurately to avoid additional tendon friction and wear. There are also not-so-obvious benefits of using our TPU tendons: They're much wider and softer than strings or cables (similar to rubber belts with a smooth surface), making them less likely to cut into the rigid finger links. Hence, we also eliminated all routing pins that are usually featured in tendon-driven hands and allowed the TPU tendons to directly contact the PLA parts, further reducing cost as well as assembly and modification effort. Secondly, these TPU tendons are also capable of elastic stretching under sudden large forces (e.g. during unplanned collisions), and can significantly reduce the risk of catastrophic failures caused by snapping of traditional tendon strings. However, these benefits also come with an increased risk of tendon stretching and wear and tear over time (PLA is not as smooth as metal routing pins after all), which were investigated through stress testing. The unexpectedly high durability of this design is presented in the next section.

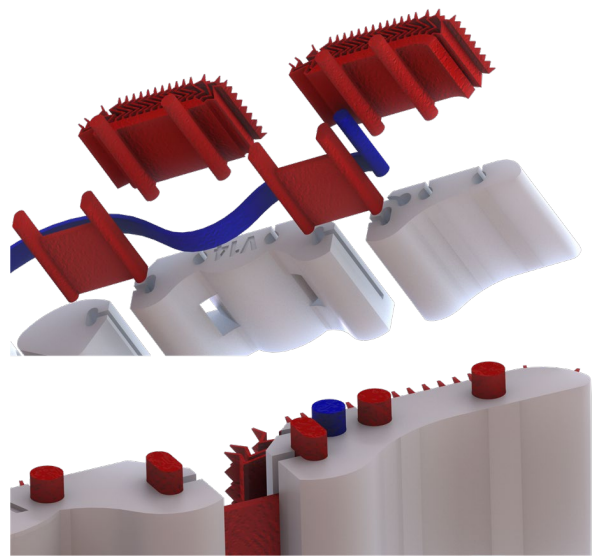

Figure 5: Finger pads, flexure joints, and tendons are all 3D printed with TPU and can be easily pushed (or hammered if required) into the PLA slots (Top). About 3 mm of the pins extend beyond the PLA surface after fitting to allow easy removal (Bottom).

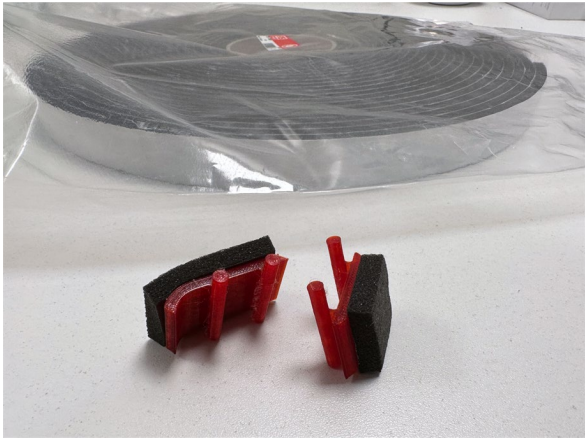

Figure 6: Finger pad adapters with self-adhesive sponge rubber strip attached.

The TPU tendons have a width of 6 mm and a thickness of 1.5 mm, whilst the flexures have a length of 16 mm, width of 20mm, and a thickness of 1.5 mm.

The Model T42 and other hands featured in the Yale OpenHand Project use two different types of cast urethane for finger pads and flexure joints. The InstaGrasp however uses the same TPU material for tendons, finger pads and flexure joints to maximize ease of fabrication. Nevertheless, the modular nature of this hand allows researchers to easily make their own improved parts to mount on the finger. In Fig. 6 for example, we showcase a TPU finger pad adapter with low-cost self-adhesive 6mm thick foam tape (£1 per meter from *RS Components UK*, item number 205-0883), which is shown to improve the grasping performance (see Section IV). If desired, it is also possible to cast urethane onto custom finger pad adapters, which should provide similar gripping performance as the T42's finger pads. Similarly, enhancements of the TPU joints and tendons are also possible but have not yet been explored yet due to the current prototype iteration being fully suitable for grasping.

It should be noted that the Amazon basics TPU does not have an official shore hardness rating from the manufacturer, but the authors of this work estimate it to be around 90A to 95A, similar to other common low-cost TPU filaments. This classifies this TPU as "hard" but enables it to be printed on low-cost 3D printers with Bowden-type extruders such as most Creality and Anycubic printers. Although softer TPU filaments of shore hardness rating 85A are less common, they can also be found on Amazon.co.uk and used to make grippier finger pads if desired. But it is recommended to use a printer with a direct-drive extruder (such as the Prusa printers) for these filaments to avoid unreliable filament extrusion.

To create a relatively soft and grippy surface using the hard A95 TPU, the finger pads are designed as an ensemble of individual 0.4 millimeter thin "sheets" whose cross section resemble an array of springs (most clearly visible in Fig. 4).

### III. STRESS TESTING

The InstaGrasp features novel components, mainly a 3D printed TPU tendon directly contacting the rigid PLA parts as well as the easy-release push-fit system. A major concern during the design stage was the reliability of the flexible TPU elements, particularly in the given configuration. Although part replacements are quick and easy, functionality and patience would suffer if parts needed to be replaced too frequently. Hence, a stress test was designed that repeatedly curls and releases one of the InstaGrasp's finger to investigate the limits of this design.

#### A. Stress Test Setup

For this stress test, a Dynamixel XM430-W250-T servo was used, given its superior construction to the XL430 (with metal gears instead of plastic ones), which we considered useful for the anticipated high number of cycles. A durability testing rig (Fig. 7) was 3D printed in PLA and directly fastened onto the servo motor. This rig features two micro switches, which are designed to be triggered by the proximal and distal links of the finger during finger tendon contraction. Note that the finger pads are not a subject of this investigation, hence earlier prototype finger pad iterations were selected based on how securely the micro switches were triggered.

As can be seen in Fig. 7B and 7C, when the finger is actuated from a relaxed state, the proximal link will first hit switch 1, followed by the distal link triggering switch 2 upon further actuation. This is designed to closely replicate an adaptive grasp. The servo is driven in extended position mode, allowing more than one rotation in case of extensive tendon stretching.

The stress test procedure is as follows: Starting from the relaxed state (defined as 0 degrees), the servo starts to rotate at approximately 57 revolutions per minute to pull the tendon. When switch 1 is triggered, the servo position and time are recorded whilst the servo continues rotating. When switch 2 is triggered, servo position and time are recorded again, and the servo is commanded to rotate in the opposite direction at the same velocity. The servo stops rotating when the 0-degree position is reached. After 1.5 seconds have passed since switch 2 is triggered (allowing ample time for full extension of the finger), the same procedure is repeated.

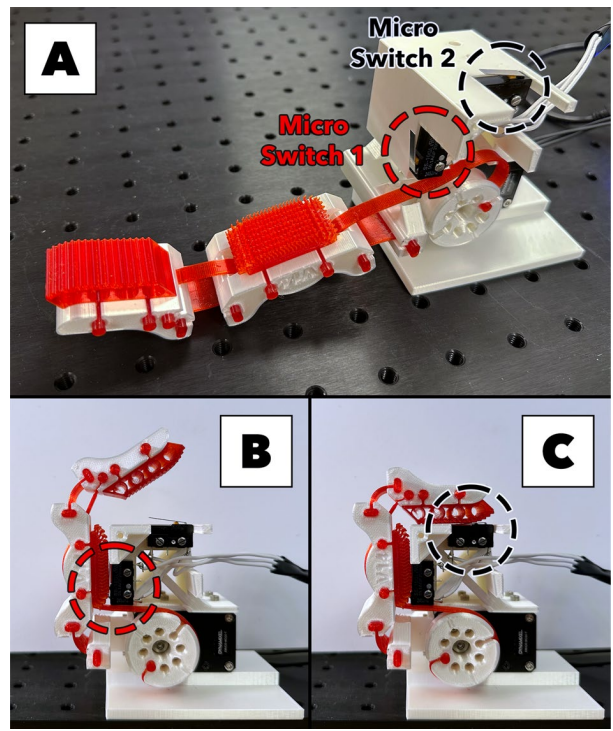

Figure 7: Durability test rig. A: Finger is relaxed, B: engaging the proximal switch, C: engaging proximal and distal switches. The tendon and joints featured in this figure have undergone more than 85,000 full curls but are still fully functional.

A full curl from the relaxed state takes around 0.8 seconds, adding up to approximately 2.3 seconds for a full curl cycle. The servo is automatically turned off and the experiment is stopped when neither switch has been triggered for more than 10 seconds.

*B. Stress Test Results and Discussions*

The stress test lasted for 86,575 full finger curls (with both switches consistently activated). All finger components managed to stay intact and functional. The experiment ended after the Dynamixel servo's metal gears got jammed, perhaps due to heat expansion after running continuously for almost 55 hours. We decided not to run the experiment again as it seemed we were now at the limit of the actuator's operation, rather than the gripper.

The recorded servo positions are plotted in Fig. 8. It is noted that the positions at proximal switch 1 are much more stable than those at distal switch 2. We hypothesize that this is due to a shorter transmission distance and less friction along the way. No tendon stretching was noticeable after the stress test, implying that the slight increase in servo position over time for proximal switch 1 is mainly due to the distal joint weakening and slowly losing its ability to return to a straight alignment.

Note that servo angles do not equate to finger link angles. However, by simplifying the flexure joints as revolute joints and considering that the moment arm of the servo pulley is roughly equivalent to those of the joints (15 mm), we can assume that a one-degree deviation in servo position corresponds to around the same in distal link angle.

Fig. 8 also shows 4 distinguishable regions. Region A (up to around 4,500 curls) is most likely a "break in" region – tendons were 3D printed in a straight line and are relatively stiff compared to string-based tendons. Thus, many curling repetitions are needed until the tendon tightly wraps around the servo pulley near the tendon-pulley connection. The next iteration of this prototype will feature tendons printed with curves and angles to take this into account. Region B is a stable region, with fluctuations within roughly 8 degrees for the distal switch and 5 degrees for the proximal switch. In region C, the distal switch's recorded positions form a peak over 6,000 repetitions. We hypothesize that this could be caused by the tendons abrading the PLA finger links thus increasing the surface friction, followed by further abrasions smoothing down the surface again. PLA powder was visible after the stress test, but not around the 20,000-repetition mark. Finally, region D is a stable region, similar to region B.

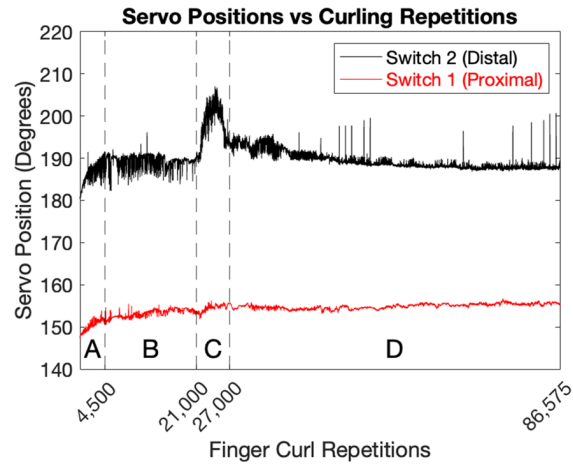

Figure 8: Recorded servo positions at both micro switches for 86,575 repetitions, with distinct regions A, B, C, and D labelled. The servo position at the relaxed finger state is defined as 0 degrees, with larger angles correlating to larger actuation.

These results show that our TPU tendons and joints are unexpectedly long-lasting, in this sense outperforming the PLA finger joints. It is recommended that users of this hand replace the PLA parts after around 20,000 uses to avoid straining the actuators with added friction. Considering that it would take more than a year to reach this number of uses if one would consistently perform 50 grasps a day, we are more than confident that this design is durable enough given the ease of part fabrication and replacements. As a rule of thumb, it would be sensible to replace the PLA finger links at least once a year. In terms of actuation repeatability and accuracy, we believe that the TPU tendons and joints are sufficient for open-loop grasping tasks and closed-loop in-hand manipulation tasks. But the current prototype iteration would not yet allow for precise open-loop in-hand manipulation.

## IV. GRASPING PERFORMANCE

In this section, the grasping performance of this gripper is assessed via a grasping experiment using objects from the Yale-CMU-Berkeley (YCB) Object and Model Set [9] shown in Fig. 9.

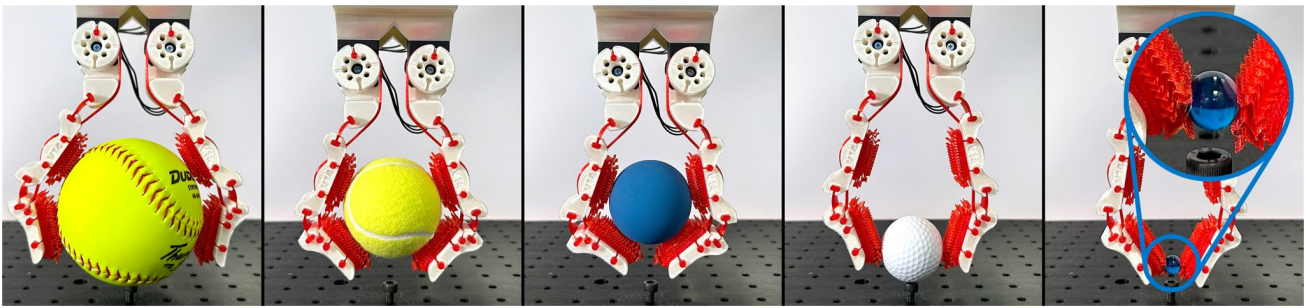

Figure 9: The five round objects used in the grasping experiments are grasped by the InstaGrasp. From left to right: softball, tennis ball, racquet ball, golf ball, and small marble.

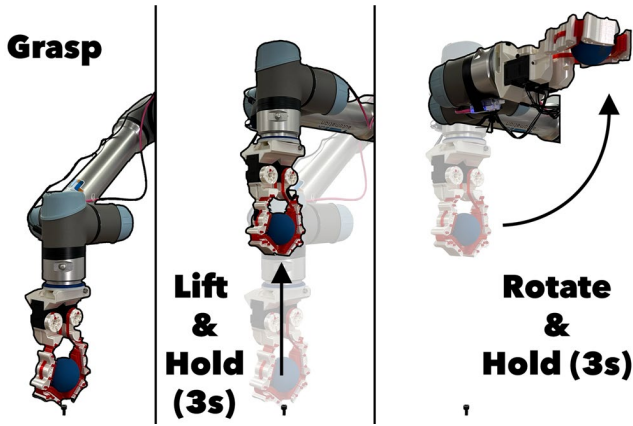

Figure 10: Grasping experiment procedure. Each object is first grasped, lifted and held for 3 seconds, and finally rotated and held for another 3 seconds.

## A. Grasping Experiment Setup

The Gripper Assessment Protocol and Benchmark from the aforementioned YCB Object Set is followed in this experiment with minor adjustments. Inspired by [8], five round objects of varying sizes, weights, and surface materials are selected from the YCB Object Set, including a softball, tennis ball, racquet ball, golf ball, and a small marble. The first three listed objects are large enough to be power grasped, whilst a pinch grasp is performed on the remaining two (golf ball and marble). Fig. 9 presents these five objects after the corresponding grasps have been performed by the InstaGrasp mounted on a UR5e robot arm from Universal Robots. All objects are initially resting on top of a cap head screw mounted on a solid aluminum optical breadboard from Thorlabs, assuring accurate positioning.

Each object is first grasped from above, lifted by 20 centimeters, then held for 3 seconds. The UR5 robot arm then rotates the gripper by 90 degrees to provide a perturbation with respect to gravity. This procedure is visually outlined in Fig. 10. Both servos are manually actuated at the same time in position mode.

This experiment is split into three tasks following the benchmarking protocol: centered grasping, x-offset grasping, and y-offset grasping. For centered grasping, the object centers are horizontally aligned with the gripper center. For the two off-centered grasping tasks (Fig. 11), the object is offset from the gripper center by 1 centimeter in either x or y direction. Each of the three tasks follows the same aforementioned grasping procedure. Note that when the hand is rotated to perform a horizontal grasp (Fig. 10 right) during the y-offset task, the object center is below the hand center. The hand uses friction in this position to hold on to the object (force closure) rather than relying on caging (form closure).

5 grasps were performed on each object for each task, totaling 75 grasps. Grasp successes (and failures) were recorded for horizontal and vertical grasps separately.

## B. Grasping Experiment Results and Discussions

Table 2 presents the number of successful grasps out of 5 for each object and task. Cells are shaded in a confusion matrix style for clarity, with darker greens indicating a higher success rate. Some objects were impossible to grasp with the default TPU finger pads but were possible with the finger pad adapters and self-adhesive sponge rubber strips attached (introduced in section II.C and shown in Fig. 5), which are shaded in blue.

Focusing on the easiest center grasping task, it is clear that vertical grasps can be reliably performed for both power grasps (softball, tennis ball, and racquet ball) as well as pinch grasps (golf ball and marble). The horizontal grasp however suffered under the insufficient friction provided by the TPU finger pads. The softball was slightly too large to hold in a comfortable power grasp (see Fig. 9), whilst the golf ball is too smooth to horizontally hold in a pinch grasp. Nevertheless, using the finger pad adapters with rubber strips attached, both softball and golf ball can be reliably grasped.

The x-offset tasks produced similar results, except for the marble which failed every time. As both fingers were actuated at the same time with the velocity, one finger consistently contacted the marble first, knocking it off the screw head it rested on. We believe that this could be overcome by actuating the fingers at separate times or velocities.

The y-offset task is the most difficult task, as objects are susceptive to ejection in the y-direction (obvious in Fig. 11) and has a smaller finger-object contact area. Both golf ball and marble could not be pinch grasped due to their small size and hence insufficient contact area. This exposes a design weakness of the InstaGrasp – although fingers are 30 millimeters wide, the finger pads only span the middle 20 millimeters. A y-offset of 10 millimeters puts the center of objects right at the edge of the finger pads. In the next iteration

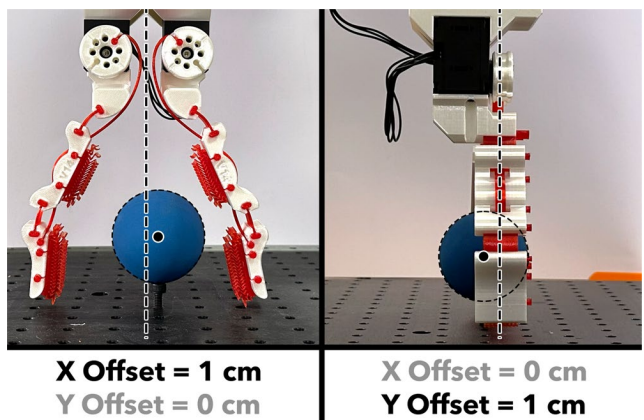

Figure 11: Initial object positions for the two off-center grasping tasks. X-offset Task (left) and y-offset task (right).

TABLE II. SUCCESSFUL GRASPS OUT OF 5

| Objects | Centered | | X-Offset | | Y-Offset | |
|---|---|---|---|---|---|---|
| | Vert. | Hori. | Vert. | Hori. | Vert. | Hori. |
| Softball (⌀ 96 mm) | 5 | 0* | 5 | 0* | 3 | 0* |
| Tennis Ball (⌀ 64.7 mm) | 5 | 5 | 5 | 5 | 3 | 0* |
| Racquet Ball (⌀ 55.3 mm) | 5 | 5 | 5 | 5 | 5 | 5 |
| Golf Ball (⌀ 42.7 mm) | 5 | 0* | 5 | 0* | 0 | 0 |
| Marble (⌀ 16 mm) | 4 | 4 | 0 | 0 | 0 | 0 |

*Can be grasped using self-adhesive foam on finger pad adapters

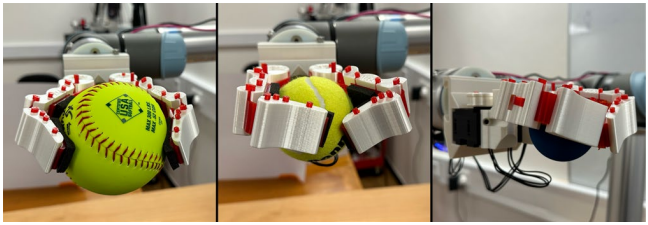

Figure 12: The InstaGrasp conforms to the object shape via twisting of flexure joints during off-center power grasping of round objects.

of this gripper, finger pads will be widened to 30 millimeters to allow for more object y-offset. On the other hand, this difficult y-offset task also highlights one of the advantages of the InstaGrasp: the TPU flexure joints are to some degree capable of twisting and are thus able to conform to the object shape with an added degree of freedom compared to revolute joints. The softball and tennis ball can be successfully grasped with soft rubber finger pads, and the racquet ball with the default TPU finger pads, even in the more difficult horizontal grasp orientation as shown in Fig. 12.

### C. Grasping Everyday Objects

This subsection showcases the InstaGrasp's ability to grasp a variety of everyday objects via different approaches. Cups (Fig. 13A) and cardboard boxes (Fig. 13B) can be power grasped; small flat objects (Fig. 13C) can be pinched; the shape-conforming ability shown in the previous subsection can be used to grasp large irregularly shaped objects (Fig. 13D); and relatively thin cylindric-like objects such as screwdrivers can be tightly grasped via actuating one servo motor slightly delayed from the other (Fig. 13E).

## V. CONCLUSION

In conclusion, the stress test and the grasping experiments indicate that the InstaGrasp is suitable for manipulation research purposes. As such, we believe that we have created a gripper design that can benefit both hand designers (who can easily use the InstaGrasp as a platform to customize) as well as researchers who simply want a capable low-cost adaptive gripper for manipulation research. However, the experiments also exposed possible design improvements, including a new tendon that is printed in a non-straight form, making it more optimised for wrapping tightly around the pulleys. Wider finger pads would also be a simple improvement.

On publication of this manuscript, we plan to open source the current design and a design featuring these improvements. To further validate the performance of this gripper, our next step is to investigate the in-hand manipulation capabilities of the InstaGrasp, via mechanics-focused methods used on the OpenHand T42 [10].

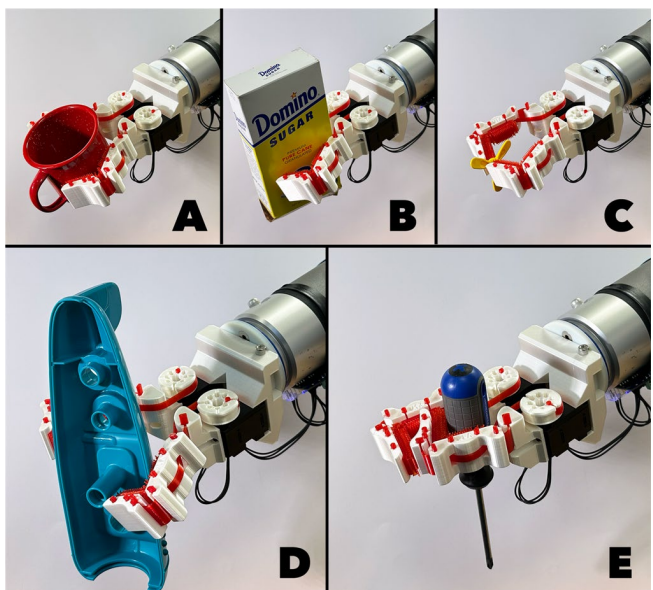

Figure 13: The InstaGrasp is capable of grasping a variety of everyday objects.